\documentclass{article}
\usepackage{arxiv}

\usepackage[utf8]{inputenc} % allow utf-8 input
\usepackage[T1]{fontenc}    % use 8-bit T1 fonts
\usepackage{url}            % simple URL typesetting
\usepackage{booktabs}       % professional-quality tables
\usepackage{amsfonts}       % blackboard math symbols
\usepackage{nicefrac}       % compact symbols for 1/2, etc.
\usepackage{microtype}      % microtypography
\usepackage{lipsum}
\usepackage{cellspace}
\usepackage{alltt}
\usepackage{calc}
\usepackage{mathtools}
\usepackage{booktabs}
\usepackage{url}
\usepackage{caption}
\usepackage{longtable}
\usepackage{amsmath}
\usepackage{amsthm}
\usepackage{amssymb}
\usepackage{bm}
\usepackage{adjustbox}
\usepackage{wrapfig}
\usepackage{epstopdf}
\usepackage{tikz}
\usepackage{framed}
\usepackage{thmtools,thm-restate}
\usepackage{enumitem}
\usepackage{algorithmic}
\usepackage{algorithm}
\usepackage{soul}
\usepackage{tcolorbox}
\usepackage{multirow}
\usepackage{array}
\usepackage{makecell}
\usepackage[colorlinks=true,breaklinks]{hyperref}

\usepackage{cleveref}

\usetikzlibrary{arrows}
\usetikzlibrary{shapes,backgrounds}

\graphicspath{{./Images/}}

\newtheorem*{theorem*}{Theorem}

\newtheorem*{lemma*}{Lemma}

\makeatletter
\renewcommand{\th@definition}{%
  \normalfont
  \thm@preskip-2 \relax
  \thm@postskip-2 \relax
}
\makeatother

\title{Modular Neural Network Approaches for Surgical Image Recognition}

\author{Nosseiba Ben Salem$^{1}$, Youn{\`e}s Bennani$^{1}$, Joseph Karkazan$^{2}$, Abir Barbara$^{1}$, Charles Dacheux$^{2}$, Thomas Gregory$^{2}$\\
  $^{1}$Universit\'e Sorbonne Paris Nord\\
  LIPN, CNRS UMR 7030 \\
La Maison des Sciences Numeriques\\
F-93430,\\ Villetaneuse - France\\
  \url{name.surname@sorbonne-paris-nord.fr}\\
$^{2}$APHP-Hôpital Avicenne, Bobigny - France\\
  \url{firstname.name@aphp.fr}
  }

\begin{document}

\maketitle

\begin{abstract}

Deep learning-based applications have seen a lot of success in recent years. Text, audio, image, and video have all been explored with great success using deep learning approaches. The use of convolutional neural networks (CNN) in computer vision, in particular, has yielded reliable results. In order to achieve these results, a large amount of data is required. However, the dataset cannot always be accessible. Moreover, annotating data can be difficult and time-consuming. Self-training is a semi-supervised approach that managed to alleviate this problem and achieve state-of-the-art performances. Theoretical analysis even proved that it may result in a better generalization than a normal classifier.
Another problem neural networks can face is the increasing complexity of modern problems,
requiring a high computational and storage cost.
One way to mitigate this issue, a strategy that has been inspired by human cognition known as modular learning, can be employed. The principle of the approach is to decompose a complex problem into simpler sub-tasks. This approach has several advantages, including faster learning, better generalization, and enables interpretability.

In the first part of this paper, we introduce and evaluate different architectures of  modular learning for Dorsal Capsulo-Scapholunate Septum (DCSS) instability classification. Our experiments have shown that modular learning improves performances compared to non-modular systems. Moreover, we found that weighted modular, that is to weight the output using the probabilities from the gating module, achieved an almost perfect classification.
In the second part, we present our approach for data labeling and segmentation with self-training applied on shoulder arthroscopy images.

\end{abstract}


\section{Introduction}
Cognitive neuropsychology has been a great inspiration and the key behind the success of data science. Our brains are the most powerful machines, that’s why they have been trying to replicate how it works and how it solves a problem into machines. 
This was the main reason to the idea of neural networks and their ability to handle complex problems. Continuing to explore the human mind, some researchers focused their attention on how our brain is subdivided into independent components. Each subset of neurons specializes in a function, and they interact with each other \cite{pshycology}. Thereby, the beginning of using modular architecture in learning. The main idea is to decompose a complex task into simple independent sub-tasks. 
It was first introduced in \cite{oldest} when they demonstrated that dividing processing produces better performance under certain conditions. Later, other researchers revealed that adding a priori knowledge helps the model  improve the results compared to non-modular. And in  recent research \cite{bengio} stated that modularity leads to  better generalization, faster learning, possesses scaling properties, and allows interpretability. Therefore, the rising interest among researchers. The modular architecture is a concept, therefore, it exists several strategies to decompose a problem. For instance, models can compete with each other to complete the same task \cite{ensemble}, or they can cooperate \cite{hierarchical}, etc. However, it may not always be evident to determine how and where to subdivide. 
\\
\\Another concept that has shown remarkable results, is the semi-supervised algorithm self-training. With the urging need for a massive number of data to train models and the difficulty or high cost of labeling data, semi-supervised learning became an active research discipline. Self-training alleviates this problem by relying on a small subset of labeled samples and a large subset of unlabeled samples, and still achieves state-of-the-art results \cite{introsl}. It assumes that the decision boundary should lie in a low-density region to obtain better generalization \cite{chapelle}. Pseudo-labeling, that is annotating the unlabeled samples iteratively, we call them pseudo-labels, and adding them to the training subset, is a common use of self-training. 
\\
\\In this work, we built systems using modular learning and self-training and applied them in two contexts. The first application aims to classify instability in a wrist structure named Dorsal Capsulo-Scapholunate Septum (DCSS). We tested different strategies of modular learning and compared them to the non-modular architecture. 
In the second application, we tested pseudo-labeling on shoulder arthroscopy images to detect structures using image segmentation. 
\\
\\To this end, we describe a brief overview of the background and investigate the related works on both modular learning and self-training in the first section. Then in the second section, we present our proposed approach for classifying DCSS using modular learning  and display the obtained results. Finally, we present the self-training approach for data labeling and segmentation.

\section{Fundamental background of the proposed approach and
related work}

\noindent In this section, we will introduce some basic background on modular learning and self-training methods and give an overview on the previous approaches in the literature.
\\% needed in second column of first page if using \IEEEpubid
%\IEEEpubidadjcol
\subsection{Modular learning}
According to neuroscience, our brain  is an amalgamation of several modules operating independently and they cooperate and communicate with each other \cite{brain}. This decomposition, inspired researchers to build a system composed of  different simple modules and then integrate their outputs. Each module is a neural network with learnable parameters and output a probability of prediction.  Modules are trained independently and are controlled by a gating module. The gating module is responsible to activate modules according to a certain rule, either set by the user or found automatically. There exist several ways to set the interaction between the different modules. For instance, one can fragment a complex problem into smaller sub-tasks. Each module is responsible for a fragment, and they cooperate between each other to create a complementary network. Or modules can compete between each other by learning to solve the same problem. 
\\
\\Using modular learning, offers major advantages  comparing to non-modular system \cite{bengio}. For instance, using multiple simple networks might significantly reduce the computational time, usually needed to train a complex model. It has been shown that modular systems improves substantially the robustness of neural networks\cite{robust}.  Combining knowledge from different modules makes the model more robust. It also reduces the model's complexity, thus optimized computation time in the training process. Moreover, the modularity limits the propagation of biases, contrary to a non-modular model. 
\noindent In 1989, Rueckl, \cite{oldest} began to study the representation and location of objects of "what" and "where" to separate. They used retinal images for classification and revealed that modular learning is more efficient in computation and learned more easily compared to non-modular network under certain circumstances. 
There exists several strategies to build modular architecture depending on the type of interaction between the structures. 
For instance, in \cite{decoupled}, they use decoupled modules where data is separated and each group is assigned to a module. The final prediction is the maximum probability from all modules. In  \cite{hierarchical}, they used hierarchical network.
Others \cite{ensemble}, and \cite{ensemble2}, use ensemble networks where all the modules are identical, and they learn to perform the same task. The final prediction is a majority vote of prediction probability from all the networks.\\
\\The most common structure is the Multiple-experts network \cite{article}. It consists of gating module and experts. The gating module controls the activation of the networks and distributes the tasks. The networks do not need to be identical it can be a hybrid system with a variety of techniques \cite{Bennani}. We used multiple experts in our case. 

\subsection{Self-Training}
Supervised learning using neural networks has seen success in several applications. To ensure a better generalization, neural networks require to be trained on a huge amount of labeled data. However, labeling data can be an expensive and time-consuming task and in some cases not accessible. Thus, the rise of a domain called semi-supervised. This domain aims to use both labeled and unlabeled data from the same or similar distribution for a specific problem. There exist several semi-supervised methods. Self-training is a common semi-supervised method. Its increasing interest is due to its better performance compared to supervised learning. This result is somehow surprising, given that we don’t add any external information. Theoretical works have proven that self-training engenders a better generalization bound given certain conditions. 

Let $N$ be the number of labeled data and $M$ the number of unlabeled data. An initial model $f_{\boldsymbol{w}^{(0)}}$ is trained from $\left\{\boldsymbol{x}_{n}, y_{n}\right\}_{n=1}^{N}$. Given the unlabeled data $\left\{\widetilde{\boldsymbol{x}}_{m}\right\}_{m=1}^{M}$ The obtained model plays the role of pseudo-labeler by computing $\hat{y}_m = \hat{f}_{\boldsymbol{w}^{(0)}}(\boldsymbol{x}_{m})$. Let $k$ be the number of selected predicted labels and let $\ell$ be the number of iterations. Then $f_{\boldsymbol{w}^{(\ell)}}$ is trained from $\left\{\boldsymbol{x}_{n}, y_{n}\right\}_{n=1}^{N} \cup \left\{\boldsymbol{x}_{m}, \hat{y}_{m}\right\}_{m=1}^{k} $ and is used as pseudo-labeler for the next iteration. Reiterate until reaching stopping criteria.
The empirical risk to be minimized is as follows 
\setlength{\arraycolsep}{0.0em}
\begin{eqnarray}
\hat{R}(\boldsymbol{W})&{}={}&\frac{\lambda}{2 N} \sum_{n=1}^{N}\left(y_{n}-f\left(\boldsymbol{W} ; \boldsymbol{x}_{n}\right)\right)^{2}\nonumber\\
&&+\frac{\tilde{\lambda}}{2 M} \sum_{m=1}^{M}\left(\hat{y}_{m}-f\left(\boldsymbol{W} ; \hat{\boldsymbol{x}}_{m}\right)\right)^{2}
\end{eqnarray}
\setlength{\arraycolsep}{5pt}
where $\lambda+\tilde{\lambda}=1$.

Similar to modular training, self-training was used decades ago from 1976 to train a model without external knowledge \cite{earlysl}. And in these recent years, researchers have shown an interest in self-training given its ability to train the model with few labeled data and a lot of unlabeled ones. The strategy of choosing pseudo-labels differs from work to another. For instance, \cite{maxprob}, selected the data having the maximum prediction probability in the current model tested on denoising auto-encoder,\cite{nearestsl}  selected the nearest observations in feature space . Others \cite{prop1}, \cite{slproportion2}, selected the proportion of the most confident data instead of setting a threshold. 

The preceding research has supposed all the classes are equals and uses a constant threshold to sort out the most confident unlabeled data to add to the training set. That is what motivated the use of dynamic threshold to prevent ignoring certain unlabeled data. \cite{zhang}, suggested adapting curriculum learning in self-training and called it curriculum pseudo labeling (CPL). Curriculum learning is an approach based on training the model starting with the data easier to learn to the more complex ones. The strategy of CPL is to adapt the threshold for each class and select unlabeled data depending on the model’s learning status. They compute performances for different classes and adjust the threshold accordingly. Meaning, if the results in a certain class are not high, the threshold ought to be lower. The authors demonstrated that their method converged faster than fixed threshold methods and achieved good results on state-of-the-art benchmarks of semi-supervised learning.  Another use of curriculum labeling in \cite{bonilla} achieved significant results. They observed that the distribution of the maximum probability prediction follows the Pareto distribution, that is, a skewed distribution. The idea is to choose unlabeled data by taking a percentile of the distribution of the maximum probability for pseudo-labeled samples. 
\section{Proposed approaches: Modular learning for classification}
\subsection{Motivation}
The first purpose of this work is to classify images of the Dorsal Capsulo-Scapholunate Septum (DCSS) to identify the stage of scapholunate instability using a modular architecture. DCSS is a structure in the wrist that ensures a scapholunate articulation \cite{dcssart}. A common disease in DCSS is  scapholunate instability. It causes severe wrist dysfunction, making it difficult for the patient to perform everyday activities \cite{dcssinstab}. Identifying the scapholunate instability requires an accurate diagnosis in order to prevent it in the early phases. We used images from wrist arthroscopy and the goal is to determine whether the structure is healthy or if it exists instability in stage 1, stage 2, or stage 3. We exploited the nature of the problem to build a modular system. We can divide the problem into, first, predicting if the structure is healthy or pathological. Then, if the DCSS is pathological, we can then use a second component to discriminate between the three stage. 
\subsection{System architecture}
The system is composed of two modules; the gating module and the discriminative module. The gating module is trained to detect if the DCSS is healthy or not from the arthroscopy image. If the DCSS is healthy, we output its prediction probability $P(y=\text{'healthy'}|X)$. If the DCSS is pathological, we feed the image to the discriminative module. The latter can be designed in different ways, and the final output can be computed differently. We choose to test three architectures (modular, modular 1 Vs 1, and weighted modular). In the following, we will detail the different cases. 
Let $H$ be the event 'healthy', $\Bar{H}$ 'pathological' and $S_i$,  $i\in \{1,2,3\} $ the stages of instability.
\paragraph{Modular (All)}
In this case, the discriminative module is trained to distinguish between the three stages of instability. Given the image predicted by the gating module to be pathological, the discriminative module computes the prediction probability $ P(S_i|X,\Bar{H}), \quad i=1, \ldots, 3 $. The discriminative module's prediction is as follows:
\begin{equation}
    \operatorname{arg\,max}_{i}P(S_i|X,\Bar{H})
\end{equation}

\paragraph{Modular (1 Vs 1)}
In this case, we built three experts competing with each other. The first module is intended to discriminate between stage 1 and 2, The second for the stage 1 and 3, and, finally, one for the stage 2 and 3. The output is the maximum from each expert. The three modules computes the probabilities in equation \ref{eqmod1} with i= \{1,2\}, \{1,3\} and \{2,3\} respectively.

\begin{equation}
    P_{m}(S_i|X,\Bar{H}) \times P_{m}(\Bar{H}|X) + P_{m}(S_i|X,H) \times P_{m}(H|X).   
    \label{eqmod1}
 \end{equation}

The discriminative module's prediction is as follows:
\begin{equation}
    \operatorname{arg\,max}_{(i,k)}P_{m_k}(S_i|X), \quad i=1,\ldots,3 \quad k=1,\ldots,3. 
\end{equation}

\begin{figure*} 
\begin{center}
\includegraphics[width=4.5in]{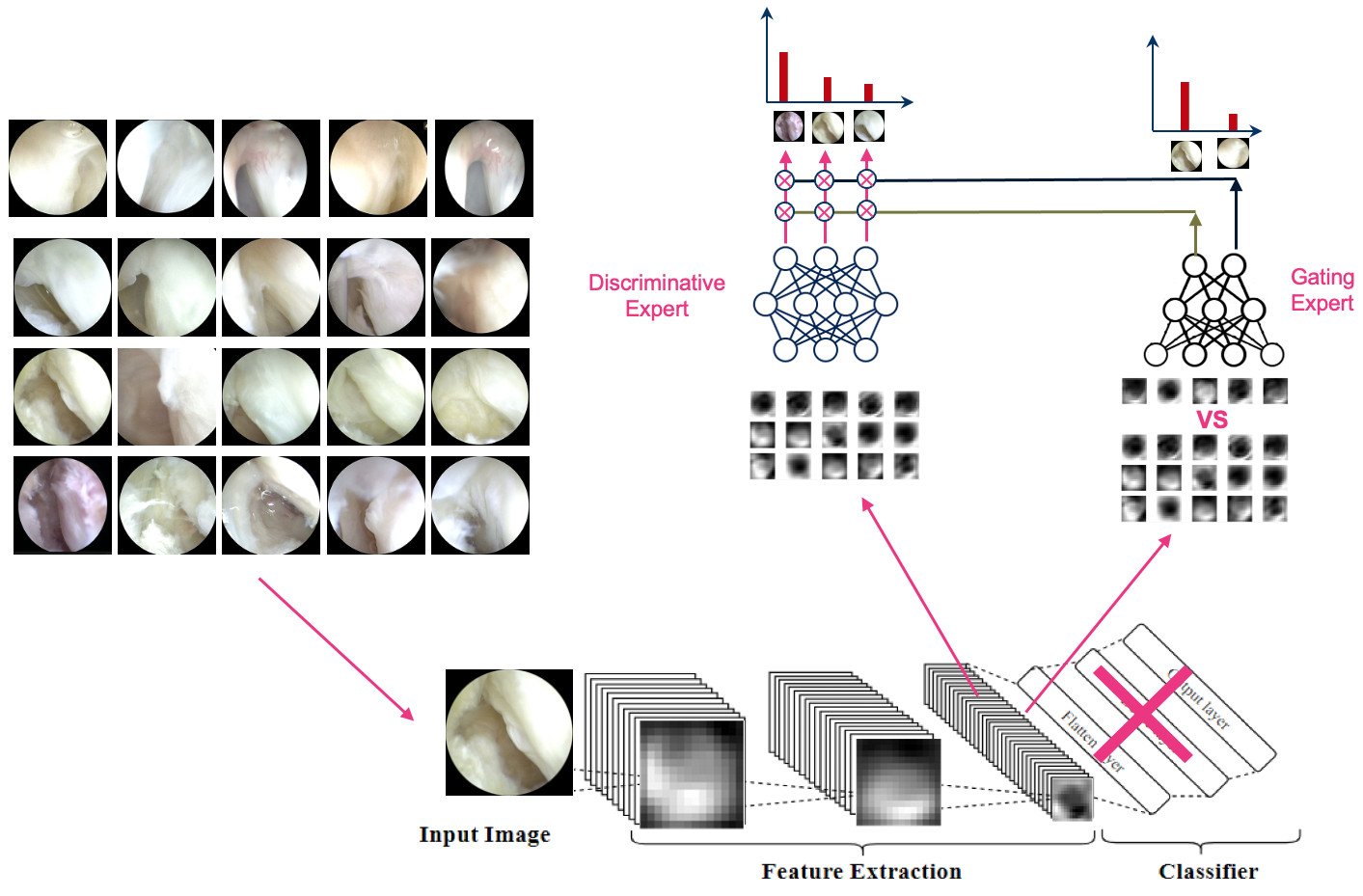}
\caption{Architecture of weighted modular learning.}
\label{fig:mel}
\end{center}
\end{figure*}

\paragraph{Weighted modular}
This system is similar to the first modular case, except for computing the output. $P(S_i|X,\Bar{H}) \times P(\Bar{H}|X) + P(S_i|X,H) \times P(H|X) , \quad i=1, \ldots, 3 $. In this case, the gating module assigns a weight to each output and picks the output that yields the highest prediction probability. The advantage of this method is to take into account the error in the gating module. In the first system, we supposed that the gating module is perfect opposing to this architecture. For instance, if the difference between the prediction probability of the gating expert is not large, it means that the module is not confident. By assigning weights to the outputs with the gating module probabilities, we add this information to the prediction. The figure \ref{fig:mel} represents the weighted modular architecture. The discriminative module's prediction is as follows:
\begin{equation}
     \operatorname{arg\,max}_iP(S_i|X), \quad i=1,\ldots,3.
\end{equation}

%\begin{multicols}{2}
%\end{multicols}
\subsubsection{Learning phase}
\begin{figure}[!t]
\centering
\includegraphics[width=3in]{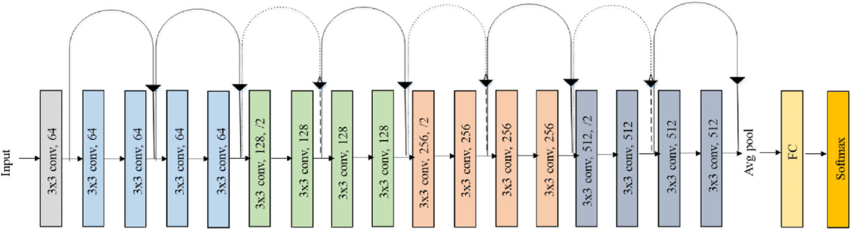}
% where an .eps filename suffix will be assumed under latex, 
% and a .pdf suffix will be assumed for pdflatex; or what has been declared
% via \DeclareGraphicsExtensions.
\caption{ResNet18 architecture\cite{resnetarch}.}
\label{fig:resnet}
\end{figure}

In each module, the network is composed of a pretrained model ResNet18 and a fully connected layer. The architecture of ResNet18 is illustrated in figure \ref{fig:resnet}. It is composed of 18 deep layers. Every four layers are similar, and each layer is composed of two blocks of deep residual network. The latter characterizes by adding an identity connection, meaning if the output of the layer is $F(x)$, the output of the block is $F(x) + x$ followed by the activation function ReLU. The residual network is an attempt to solve the problem of vanishing gradient, that is, when the gradient reaches zero quickly \cite{resnetpaper}. And the final layer consists of average pooling followed by a fully connected layer and a softmax. 

\subsection{Experiments}

\subsubsection{Dataset}
\begin{figure}[!t]
\centering
\includegraphics[width=2.5in]{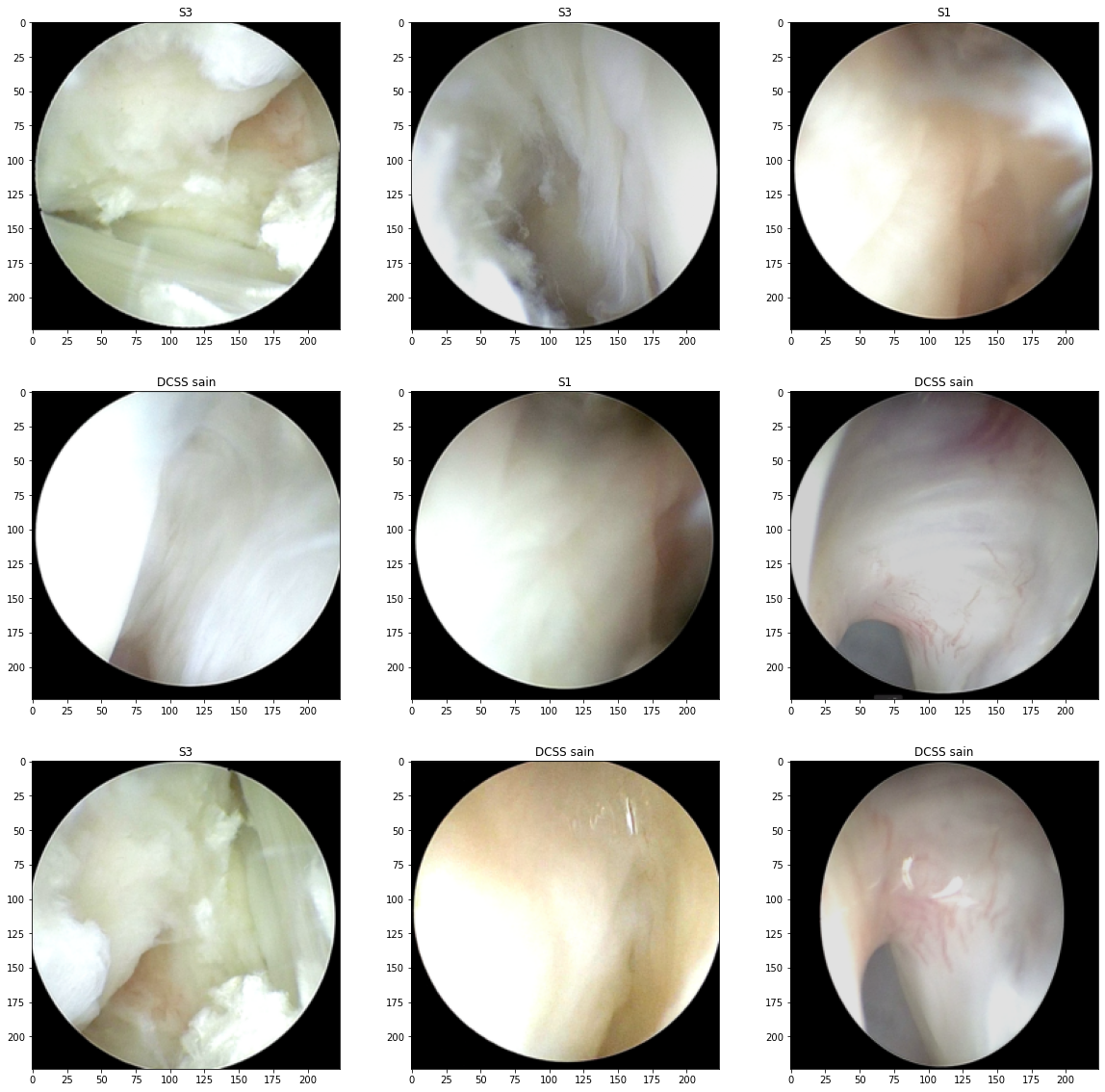}
% where an .eps filename suffix will be assumed under latex, 
% and a .pdf suffix will be assumed for pdflatex; or what has been declared
% via \DeclareGraphicsExtensions.
\caption{Visualization examples of  DCSS images.}
\label{fig:data}
\end{figure}

The images used in this work were sampled from arthroscopic videos on Dorsal Capsulo-Scapholunate Septum (DCSS) performed at Avicenne
Hospital . The collected data includes 840 photos in total, including 105 healthy DCSS images (12.5\%), 173 (20,6\%)  in stage 1, 187 (22.2\%)  in stage 2, and 375 in stage 3 (44.64\%). Entry from each patient was equally distributed.
In order to maintain a balance between the classes, we have expanded the training set using image manipulations such as random rotations, adjusting the brightness, zooming in on the image, etc. After transformation, we were able to obtain 3076 healthy DCSS images, 1111 in stage 1, 1126 in stage 2, and 1279 in stage 3. The image's size was then resized to $224\times 224$.
The figure \ref{fig:data} presents samples from the dataset.

\subsection{Setting}
We used 80\% (6592) of the dataset for training and 20\% (252) for testing. We used stochastic gradient descent to optimize the model and a learning rate of $0.001$. We trained for 10 epochs. We used the same hyperparameters for both models. The first model was trained on all the training datasets labeled healthy, not healthy and the second model was trained on only non-healthy DCSS. 
\subsubsection{Results and discussion}
We evaluated the three types of modular learning using accuracy with an estimation of the confidence interval. The confidence interval was computed as explained in \cite{Bennani}. They assume that the probability distribution of the number of errors is binomial, so we calculate the interval according to the formula below
\begin{equation}
    I_\alpha=\frac{P+\frac{Z_\alpha^2}{2 N} \pm Z_\alpha \sqrt{\frac{P(1-P)}{N}+\frac{Z_\alpha^2}{4 N^2}}}{1+\frac{Z_\alpha^2}{N}},
\end{equation}

where $N$ is the number of observations in the test set, $P$ is the probability that an observation is an error, and, $\alpha\%$ is the confidence level, usually set to $95\%$, $Z_\alpha$ is a value given by the normalized centered Gaussian distribution ($Z_{95 \%}=1.96$  and $Z_{90 \%}=2.48$). \\
\\The table \ref{tab:dcss} summarizes the results obtained. As we can see in the table, modular learning increased the accuracy from $97.6\%$ to $98.4\%$ and a tighter confidence interval of $[97.79\%; 100\%]$ with an almost perfect gating module. These results demonstrate that adding a priori knowledge from the modular nature of the problem allowed a better generalization and a finer decision boundary than the non-modular architecture. The decomposition of the task makes the problem less complex, thus a bigger probability to converge towards the ground truth function. 

\begin{figure}[!t]
\centering
\includegraphics[width=2.5in]{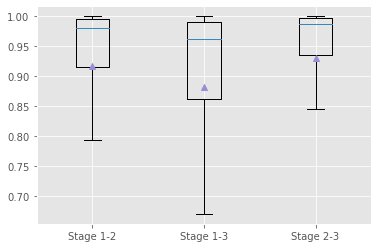}
% where an .eps filename suffix will be assumed under latex, 
% and a .pdf suffix will be assumed for pdflatex; or what has been declared
% via \DeclareGraphicsExtensions.
\caption{Distribution of difference in prediction.}
\label{fig:diff}
\end{figure}

\begin{figure}[!t]
\centering
\includegraphics[width=3.5in]{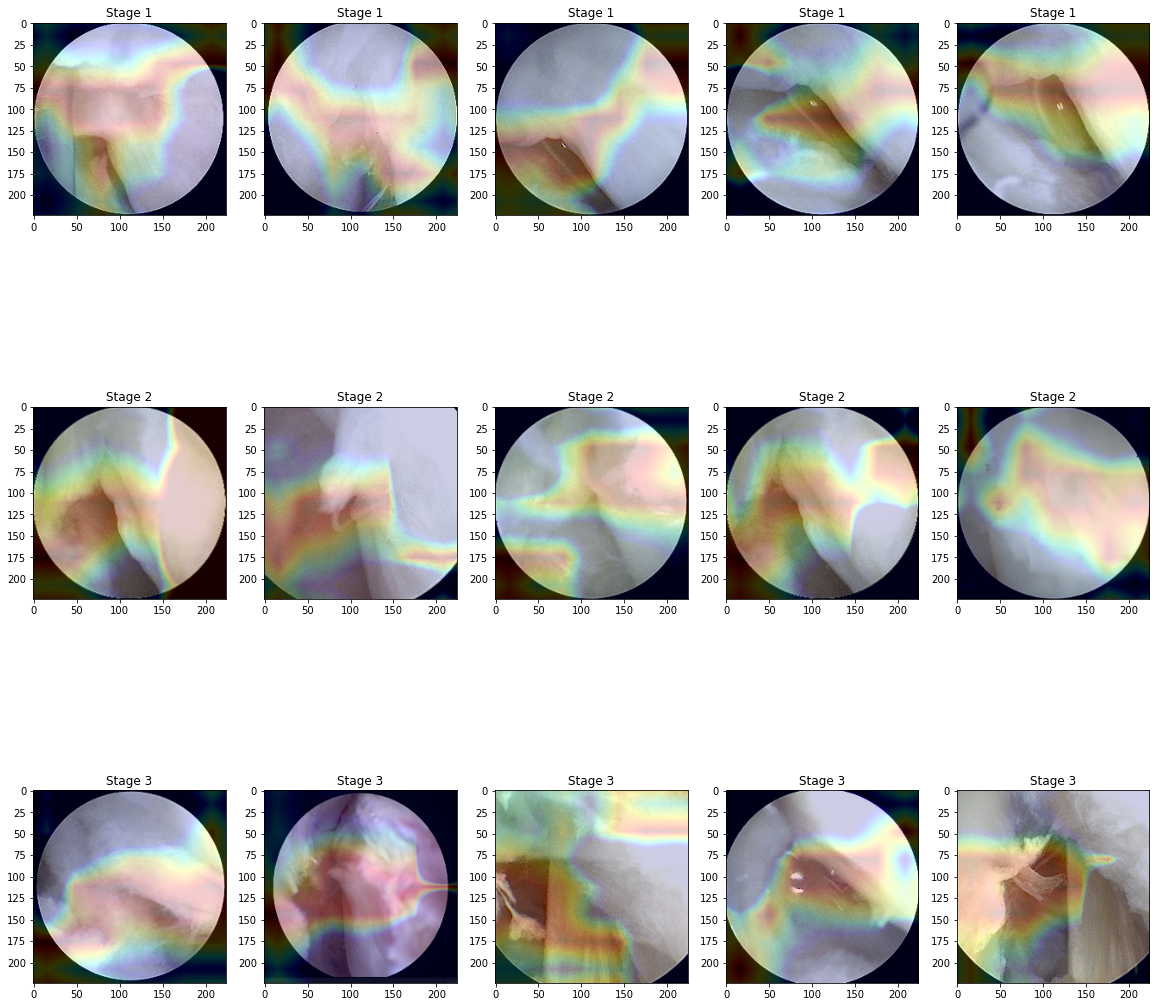}
% where an .eps filename suffix will be assumed under latex, 
% and a .pdf suffix will be assumed for pdflatex; or what has been declared
% via \DeclareGraphicsExtensions.
\caption{Interpretation of model using GradCam.}
\label{fig:grad_2}
\end{figure}

\begin{figure}[!t]
\centering
\includegraphics[width=3.2in]{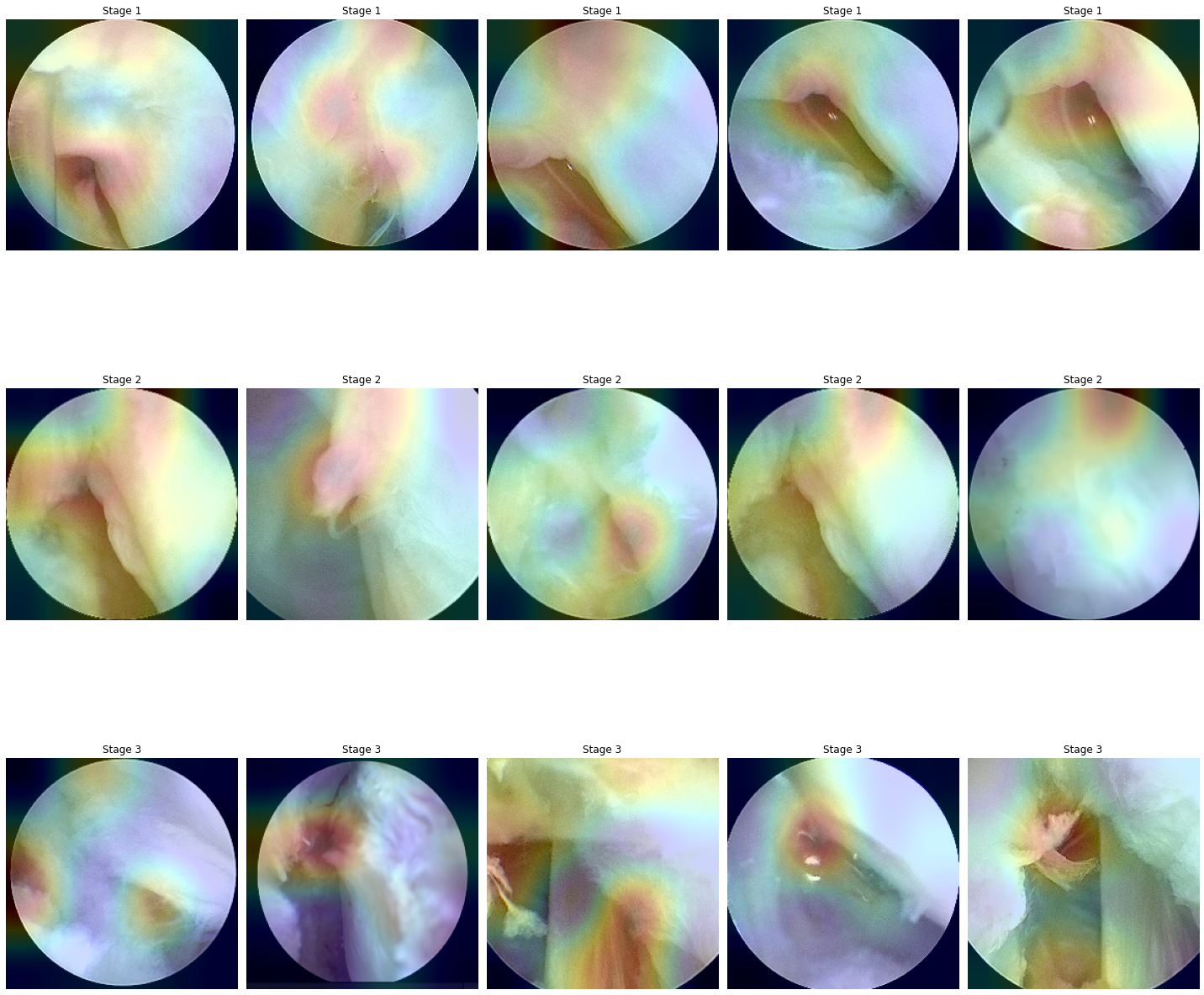}
% where an .eps filename suffix will be assumed under latex, 
% and a .pdf suffix will be assumed for pdflatex; or what has been declared
% via \DeclareGraphicsExtensions.
\caption{Interpretation of model using Sobol attribution method.}
\label{fig:sobol_2}
\end{figure}

The experiments in \ref{tab:dcss}, \ref{tab:matrix} also showed that weighted modular had better performance than the first case modular architecture with $98.81 \%$ accuracy. This indicates that adding probabilities from the gating module calibrates the results. \\
\\We investigated the distribution of the difference between the prediction probability of the different classes in the discriminative module. We can see in figure \ref{fig:diff} that the median difference is almost one, which proves that the discriminative module is confident in its decision. We can also notice a high difference between the prediction probability of stage 2 and 3 meaning that the model is able to distinguish between the two stages confidently. However surprisingly, the probabilities between stages 1 and 3 are relatively small. By analyzing the images further, we notice the presence of redness in both stages, which could be the reason for such results. \\
\\Since the problem is medical, it is important to explain the choice made by the model. However, neural network is a black box. Therefore, we attempted to evaluate two methods to interpret the results using a heatmap on images. The  pixels with the higher contributions to the neural network's prediction are located in hotter areas. The first method is called GradCam (gradient-weighted class activation mapping), which creates visual explanations \cite{art:gradcam}. Grad-CAM creates a coarse localization map that highlights key areas in the image by using the gradients of any target concept and flowing into the final convolutional layer. This makes it possible to visualize the results of various CNN model layers. 
\\
\\The second approach used for explainability is Sobol attribution method \cite{sobol}. It relies on sensitivity analysis techniques commonly used in physics and economics, in particular the Sobol indices. The method, not only does it identify the contribution of each image regions to the neural network's decision but also, the sobol indices are able to estimate High-order interactions between the regions in the image and their contributions. To estimate the indices, the authors used Jansen estimator with Quasi-Monte Carlo for sampling. First, they sample masks from the sequence, apply them to the image with a perturbation operator, then obtain prediction scores by forwarding the obtained image to the system. The explanations are then computed using the masks and the prediction scores.
In figure \ref{fig:grad_2} and \ref{fig:sobol_2}, we evaluate displays the visual output of the discriminative model of the two methods. We notice that the GradCam provides a continuous region, and we notice a certain pattern for each class, while the Sobol attribution method gives a finer heatmap. In both cases, both methods were able to locate the rip in the DCSS in Stage 3. 

\begin{table}[!t]
%% increase table row spacing, adjust to taste
\renewcommand{\arraystretch}{1.3}
% if using array.sty, it might be a good idea to tweak the value of
% \extrarowheight as needed to properly center the text within the cells
\caption{Average accuracy on different modular architecture}
\label{tab:dcss}
\centering
%% Some packages, such as MDW tools, offer better commands for making tables
%% than the plain LaTeX2e tabular which is used here.
\begin{tabular}{|c||c|c|}
\hline
Architecture & Accuracy (\%)  & Confidence interval(\%)  \\
\hline
Non-modular
 & 97.6 & [96.52 ; 100]  \\
\hline
 Modular (All)
        &  98.4     & [97.79 ; 100]    \\
\hline
Modular (1 Vs 1) & 94.4 & [91.1 ; 100] \\
\hline
Weighted modular  & 98.81 & [98.4 ; 100] \\
\hline
\end{tabular}
\end{table}

\begin{table}[!t]
\setlength\tabcolsep{2pt}

\caption{Evaluation matrix of weighted modular on test set}
\label{tab:matrix}
\centering
\begin{tabular}{|l|l|r|r|r|r|l|}
\hline \multicolumn{2}{|l|}{ }  & \multicolumn{4}{|c|}{ Predicted result  } & Recall (\%)  \\
\cline { 3 - 6 } \multicolumn{2}{|l|}{} & Healthy & Stage 1 & Stage 2 & Stage 3  &  \\
\hline Ground truth  & Healthy & $30$ & $0$ & $0$ & $0$  & $100$  \\
\cline { 2 - 7 } & Stage 1 & $2$ & $60$ & $1$ & $0$  & $95.23$  \\
\cline { 2 - 7 } & Stage 2 & $0$ & $0$ & $62$ & $0$  & $100$  \\
\cline { 2 - 7 } & Stage 3& $0$ & $0$ & $0$ & $97$ & $100$  \\
\hline Precision (\%) & & $93.75$ & $100$ & $98.41$ & $100$ &  \\
\hline
\end{tabular}

\end{table}


\section{Proposed approaches: Self-training for data labeling and segmentation }

\subsection{Motivation}
\label{moti:1}
The main objective of this section is to, first, label arthroscopy images using self-training and then detect and identify structures from arthroscopy videos with a modular architecture.
We used videos of shoulder arthroscopy. In one second of video, we can obtain from 25 to 30 images per second. Therefore, several minutes in a video can produce several images, thus labeling these data can be difficult and time-consuming. The idea is to select few pictures from each video and apply the self-training process for pseudo-labeling, that is, producing predicted labels from a small set of labeled data and adding them to the training dataset, of the rest of the video. This way, we exploit all the available data we have.
\subsection{System architecture}
\label{sec:arch}
 The model for the self-training (i.e. pseudo-labeler) is built using a segmentation network. It takes as input an arthroscopic image and its ground truth mask with 5 classes (background, long biceps tendon, subscapularis tendon, supraspinatus tendon, cartilage (glenoid and humeral)), and it outputs a prediction mask. 
We have found while training the segmentation task, that the model was able to perform a binary segmentation, meaning identifying the structure versus the background. However, it fails to discriminate between the different classes. That's what motivates the use of modular architecture in learning. 
The first network learns to distinguish between the background and the anatomical structure. Then, we remove the predicted background from the initial image and feed it to the second network. The latter is trained to learn to discriminate between the four classes. 
\newcommand{\intervalleoo}[2]{\mathopen{]}#1\,;#2\mathclose{[}} \\
\\The self-training phase, consists in, first, training the two models with the available annotated images. Then, we predict the mask for all the images in the unlabeled set.  Afterward, we select only the images that produced a mean prediction probability superior to a threshold. We set another threshold to filter the selected outputs, pixel by pixel, according to their prediction probability. The experiments have shown that adding the second filter to the predicted images increases the accuracy. 
Let $\hat{y}_m$ be a predicted image from the model. $\hat{y}_{m_{i,j}}$ represents the pixel at position $(i,j)$. Let $T  \in \mathopen{[}0,1\mathclose{]}$ be the threshold and the Softmax function is defined as  $S: \mathbb{R}^{K} \rightarrow(0,1)^{K}$ where $K \geq 1$; 
 $S(\mathbf{x}_{i})=\frac{e^{x_{i}}}{\sum_{j=1}^{K} e^{x_{j}}} \quad$ for $i=1, \ldots, K$ and $\mathbf{x}=\left(x_{1}, \ldots, x_{K}\right) \in \mathbb{R}^{K}$. The image is filtered as follows;

\begin{equation}
     \hat{y}_{m_{i,j}}= 
\begin{cases}
   \hat{y}_{m_{i,j}},& \text{if } \operatorname{arg\,max}_{(i,j)}S(\hat{y}_{m_{i,j}})\geq T\\
    0,              & \text{otherwise.}
\end{cases}
\end{equation}

We add the final outputs to the training set and reiterate until reaching the stopping criteria; either all the images are annotated or the algorithm reached the maximum number of iteration. 

Let $P  \in \mathopen{[}0,1\mathclose{]}$ be a threshold, then in each iteration the set of selected images is as following: 
\begin{equation}
    \hat{I} \coloneqq \{ \hat{f}_{\boldsymbol{w}^{(\ell)}}(\boldsymbol{x}_{m}) \mid \operatorname{arg\,max}_{\boldsymbol{x}_{m}} S(\hat{f}_{\boldsymbol{w}^{(\ell)}}(\boldsymbol{x}_{m})) \geq P \}
\end{equation}

and the output prediction is defined as follows:
\begin{equation}
    \hat{y}_m = \delta_{\operatorname{arg\,max}_{(i,j)}S(I_{m_{i,j}}) \geq T} \hat{I}_m; \quad 0\leq m \leq n(\hat{I}), 
\end{equation}

where $n(\hat{I})$ is the cardinality of the set $\hat{I}$.
\subsubsection{Learning phase}
\begin{figure}[!ht]
    \centering
    \includegraphics[width=3.2in]{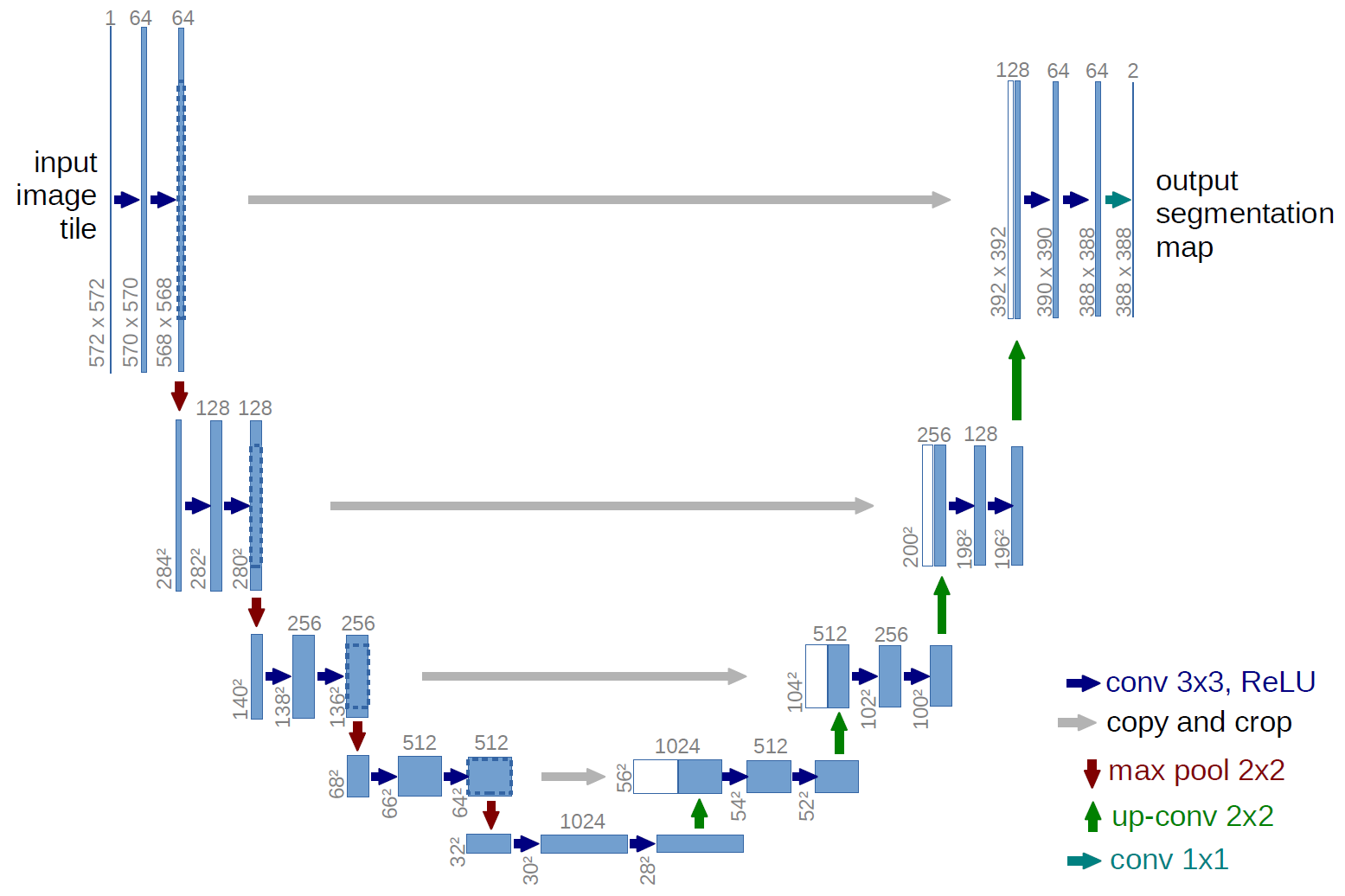}
    \caption{UNet architecture\cite{unet}.} 
    \label{fig:unet}
\end{figure}

The architecture of each module is inspired by UNet \cite{unet}. The choice of this particular pretrained model is motivated, first, by its common use in medical applications. The second key is its ability to perform well with few labeled images \cite{unet}. The architecture of UNet is U-shaped. It consists of an encoder (contracting path) and a symmetric decoder (expanding path). The contracting path consists of 2 blocks of $3\times 3$ convolution followed by the activation function rectified linear unit (ReLU) then a $2\times 2$ max pooling. These operations are repeated five times. The number of channels doubles in each time (64, 128, 256, 512, 1024). The expanding path consists of the same blocks and ReLU and uses $2\times 2$ for the up convolution. The final layer is a $1 \times 1$ convolution. The figure \ref{fig:unet} displays the UNet architecture. 

%%%%%%%%%%%%%%%%%%%%%%%%%%%%%%%%%%%%%%%%%%%%%%%%%%%%%%%%%%%%%%%%%%%%%%%%%%%%%%%%%%%%%%%%%%%%%%%%%%%%%%%%%%%%
\subsection{Experiments}
\label{sec:Experiment}
%%%%%%%%%%%%%%%%%%%%%%%%%%%%%%%%%%%%%%%%%%%%%%%%%%%%%%
\subsubsection{Dataset}
\begin{figure}[!ht]
    \centering
    \includegraphics[width=2.8in]{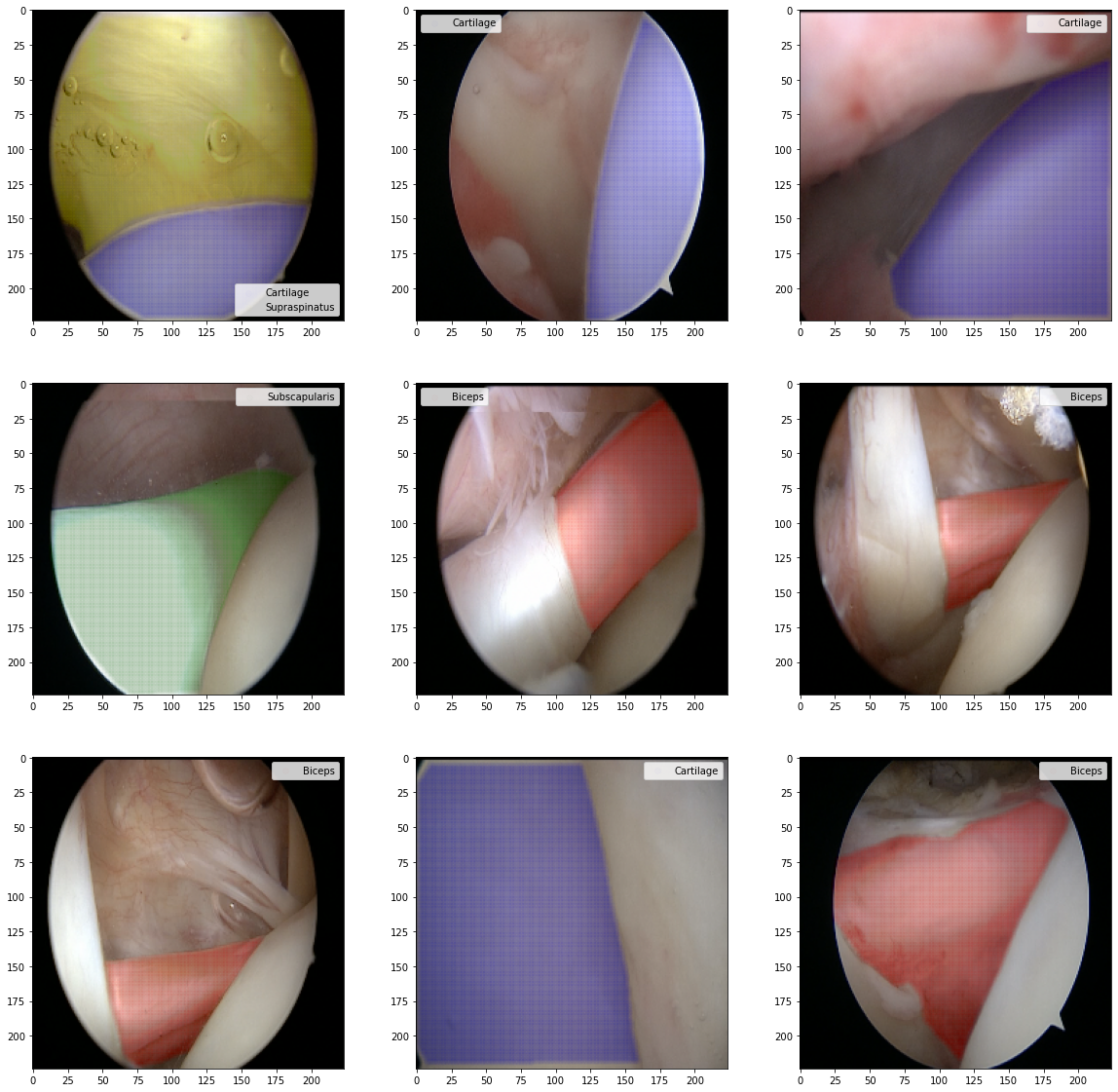}
    \caption{Visualization examples of  segmentation on shoulder arthroscopy image. Each color corresponds to the classes accordingly; red: Biceps, blue: Cartilage ,green: Subscapularis , yellow: Supraspinatus.} 
    \label{fig:data2}
\end{figure}

The data of our second study were collected by video recording of shoulder arthroscopies performed at Avicenne Hospital between November 2021 and June 2022.
The videos of arthroscopies could be retrieved directly from the arthroscopy video console in mp4 format.
From these arthroscopies, a total of 459 images were  extracted (121 of long biceps tendon, 41 of supraspinatus tendon, 181 of subscapularis tendon, and 116 of glenoid or humeral cartilage), and saved in png format. 
The images were annotated by a surgeon using the software "LabelStudio". Labeling was made using polygon annotations. The latter ensures a more precise annotation for irregular shapes. Figure \ref{fig:data2} presents examples of shoulder arthroscopy images with their segmentation. 

\subsection{Setting}
Data was split into $70\%$ train set (321 images), and $ 30 \%$ test set (138 images). We used $10 \%$ of the training set (32 images) as labeled data to train the first model in self-training and we supposed the $90\%$ of the training set (289 images) unlabeled data. The final evaluation was made on the test set. The images were resized to $224 \times 224$. We used an Adam optimizer and 15 number of epoch. After experiments, we set the threshold to $0.7$.

\subsubsection{Results and discussion}
\begin{figure}[!h]
    \centering
    \includegraphics[width=3.0in]{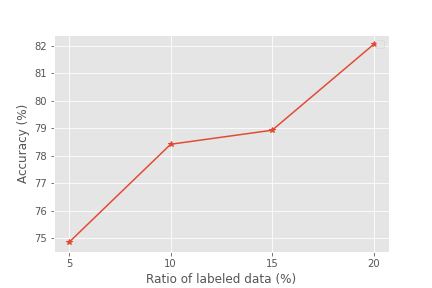}
    \caption{Comparison of ratio of labelled data with respect to accuracy.} 
    \label{fig:ratio}
\end{figure}

\begin{figure}[!h]
    \centering
    \includegraphics[width=3.5in]{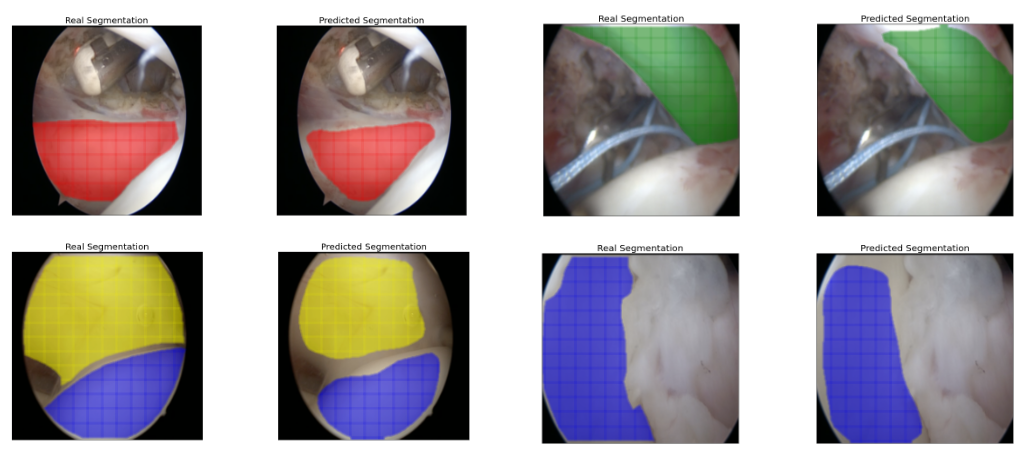}
    \caption{Examples of segmentation results predicted with self-training. Each color corresponds to the classes accordingly; red: Biceps, blue: Cartilage ,green: Subscapularis , yellow: Supraspinatus. } 
    \label{fig:preds}
\end{figure}

We use a threshold on the prediction probability to choose from the unlabeled subset, and the samples to add to the training set (improved accuracy of self-training from 76\% to 80\%). We varied the threshold to analyze its effect on performance. We reported the results on the test set.  As shown in table \ref{tab:shoulder}, increasing the threshold improves the accuracy. However, for the threshold $0.9$, the performance degrades. One reason can be that the strict selection of samples engenders the model to train on the same samples, therefore overfits. The optimal threshold value is $0.7$ with an accuracy of $80\%$. We also show that the class Supraspinatus, presented in yellow in figure \ref{fig:data2} has the highest accuracy compared to other classes. We found that adding the pixel-by-pixel threshold, explained in  \ref{sec:arch}, improved the accuracy (from $76\%$ to $80\%$). We observed the effect of the ratio of the labeled data in the first iteration on the performance in figure \ref{fig:ratio}. We notice that the greater percentage, the higher accuracy. We choose the ratio $10\%$. Samples of the segmentation predictions on the test set are shown in figure \ref{fig:preds}.

%\begin{tabular}{|l|l|r|r|r|r|l|}
%\hline \multicolumn{2}{|l|}{ }  & \multicolumn{4}{|c|}{ 
\begin{table*}[!t]
\caption{Comparison of threshold value in self-training.}
\centering
%\begin{tabular}{@{\extracolsep{\fill}} l *{5}{d{2.4}} }
%{\textwidth}
%\begin{tabular}{|l|l|r|r|r|r|l|}
\begin{tabular}{lllllll}
\toprule
 \multicolumn{2}{c}{Threshold} & \multicolumn{5}{c}{Classes} \\
 \cmidrule(r){3-7}
 & & \multicolumn{1}{c}{Mean value} & \multicolumn{1}{c}{Biceps}  & \multicolumn{1}{c}{Cartilage} & \multicolumn{1}{c}{Subscapularis}  & \multicolumn{1}{c}{Supraspinatus}\\
\midrule

0.6 &  Accuracy      & 71.2\% & 64.48\% & 37.25\%& 25.95\% & 84.35\%   \\
    & IOU          &0.31  & 0.47 & 0.29 & 0.15  & 0.34 \\
0.7 & Accuracy       & 80\% & 70.76\% &65.25\% & 64.21\%  & 85.73\%   \\
    & IOU& 0.3& 0.3 & 0.4&0.4&0.11                                               \\
              
0.8 & Accuracy & 79.55\% &44.91\% & 62.35\%  & 50.83\% & 92.32\% \\
    & IOU &  0.39 & 0.35 & 0.29 & 0.29 & 0.61                                                       \\
0.9 & Accuracy     & 76.73\% & 43.66\% & 73.17\%    &43.03\% & 88.64\% \\
    & IOU  & 0.3&0.38&0.39&0.16&0.39                                                             \\
\bottomrule         
\end{tabular}
\label{tab:shoulder}
\end{table*}

%%%%%%%%%%%%%%%%%%%%%%%%%%%%%%%%%%%%%%%%%%%%%%%%%%%%%%%%%%%%%%%%%%%%%%%%%%%%%%%%%%%%%%%%%%%%%%%%%%%%%%%%%%%%

\section{Conclusion}
In this paper, we first introduce a modular framework for Dorsal Capsulo-Scapholunate Septum classification. We have shown that decomposing the problem into a gating module and discriminative module improved the performance. Moreover, including the contribution of the gating module as weights, achieved the best results compared to the different strategies tested.  Additionally, we presented a modular self-training approach for data labeling and segmentation. Our results validated the improvement of results by using self-training. However, while modular learning outperformed non-modular systems in the first problem, it was not the case for the second problem. The reason may be related to the strategy of dividing the problem. In the classification problem, we used a priori knowledge on the classification subject, contrary to the segmentation problem. \\
\\In conclusion, modular learning is an efficient approach to reduce the complexity of a model and improve model generalization. However, it is crucial to manage when and where to divide a problem.  In future work, a different strategy should be developed and further analyzed for modular self-training segmentation. Moreover, propose a new measure of feature selection for interpretability on convolutional neural networks.

\bibliographystyle{splncs04}
\bibliography{ref} 

\end{document}